\documentclass[letterpaper,10pt,conference]{ieeeconf}
\usepackage{amsmath}
\usepackage{amssymb}
\usepackage{cite}
\usepackage{graphicx}
\usepackage{booktabs}
\usepackage{capt-of}
\usepackage{microtype}
\usepackage{etoolbox}
\IEEEoverridecommandlockouts

\title{\LARGE\bf
EchoNav: Evidential BEV Completion from Historical Observations for Visual Navigation
}
\author{Diwen Liu$^{1}$, Yiyang Shao$^{1}$, Borui Wang$^{1}$, Chi Zhang$^{2}$,\\
Jizhuo Chen$^{3}$, Jiaming Wang$^{3}$, Benyun ZHAO$^{1}$, Zhongyu Li$^{1}$, Yunhui Liu$^{1,*}$%
\thanks{\raggedright $^{1}$The Chinese University of Hong Kong, Hong Kong SAR, China.
\newline\texttt{\{stevenliudw,1155250209,1155267575,}
\texttt{benyunzhao\}@link.cuhk.edu.hk},
\newline\texttt{zhongyli@cuhk.edu.hk}, \texttt{yhliu@mae.cuhk.edu.hk}.}%
\thanks{\raggedright $^{2}$The University of Hong Kong, Hong Kong SAR, China.
\texttt{zc1q84@connect.hku.hk}.}%
\thanks{\raggedright $^{3}$National University of Singapore, Singapore.
\newline\texttt{jamesjizhuochen@gmail.com}, \texttt{jiaming@comp.nus.edu.sg}.}%
\thanks{\raggedright $^{*}$Corresponding author: Yunhui Liu (\texttt{yhliu@mae.cuhk.edu.hk}).}%
}
\IEEEaftertitletext{
\begin{minipage}{\textwidth}
\centering
\includegraphics[width=\textwidth]{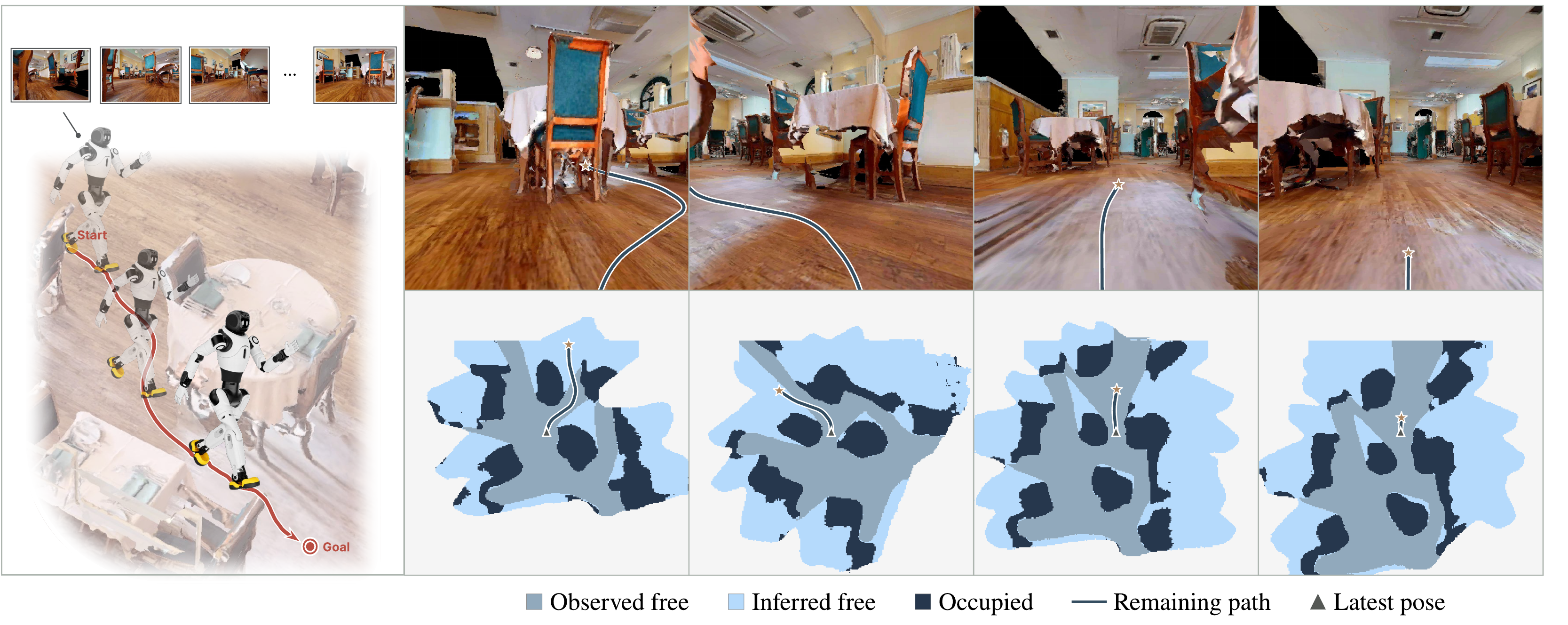}
\captionof{figure}{\textbf{EchoNav: temporal BEV mapping and evidential occupancy completion for visual navigation.} The predictor integrates RGB history to retain previously observed structure and infer hidden obstacles and free space. Paths are planned on the predicted BEV maps and projected onto RGB views. Off-white denotes regions outside the input camera-footprint union.}
\label{fig:introduction}
\vspace{8pt}
\end{minipage}
}

\begin{document}
\maketitle
\thispagestyle{empty}
\pagestyle{empty}

\begin{abstract}
Visual navigation in cluttered environments requires robots to recover previously observed structure and infer obstacles and free space that remain occluded. We present a framework that combines temporal aggregation and occupancy completion in an evidential bird's-eye-view (BEV) representation predicted from short RGB sequences. A decoder adapts frozen VGGT-Omega features and combines historical residual proposals using cellwise weights and a learned no-update option. The BEV predictor produces a local map in the latest ego frame without depth or camera-pose inputs or explicit geometric warping. The evidential model treats observed free space separately and predicts occupancy with associated confidence across the remaining supported region. We train the predictor on simulated RGB sequences paired with complete and visibility-masked occupancy annotations. Our experiments evaluate navigation in simulation and on a wheeled robot, with additional demonstrations on quadruped and humanoid robots. In simulated sequential two-goal navigation, our system achieves 86.3\% success, outperforming the strongest evaluated baseline by 32.2 percentage points. Across 60 wheeled-robot trials in 20 scenes, it achieves 73.3\% success, compared with 33.3\% for the strongest evaluated baseline.
\end{abstract}

\section{Introduction}
\label{sec:introduction}

Mobile robots navigating cluttered environments must make spatial decisions from incomplete visual observations (Fig.~\ref{fig:introduction}). Each camera view reveals only part of the surrounding layout. Relying only on the latest view can cause a robot to revisit unproductive routes or become trapped in local minima and fail to reach its goal. Observations retained within the RGB window provide evidence about nearby regions that are no longer visible in the current image. The more complete map gives the planner additional local context to choose more efficient paths. Yet some obstacles and free space remain hidden behind furniture and walls throughout this process. Predicting these hidden regions can inform planning, but the robot must account for the uncertainty of those predictions. The challenge is therefore to build a local BEV map that retains useful information across time, distinguishes observed space from inferred structure, and indicates how reliable its predictions are.

Learned occupancy anticipation has demonstrated that visual context can help predict unseen structure~\cite{ramakrishnan2020occant}. In autonomous driving, occupancy completion has shown strong results in scene understanding by predicting occupancy in both visible and occluded regions~\cite{cao2022monoscene,li2023voxformer}. For mobile robots, anticipating obstacles and free space beyond current observations can similarly provide a more complete map for planning, supporting exploration and navigation before those regions are directly observed~\cite{ramakrishnan2020occant,sharma2023proxmap}.

BEV occupancy maps place obstacles, free space, and candidate paths in a common top-down frame. This representation makes the basis of motion decisions transparent. The map describes the environment, while the planner and controller handle robot footprints, clearance margins, and motion constraints. This separation leaves the choice of planning algorithm open to the task and robot platform.

To address these challenges, we propose an RGB-based framework for temporal BEV mapping and occupancy completion. It combines evidence from earlier observations with predictions of persistently hidden obstacles and free space, producing occupancy and reliability estimates for local planning. The BEV predictor requires neither depth nor camera-pose inputs and uses no explicit geometric warping.


The main contributions are:
\begin{enumerate}
    \item \textbf{Spatially selective temporal BEV fusion.} We propose a framework that integrates RGB histories into an ego-centric occupancy BEV. Historical residual proposals from a shared decoder are weighted independently at each map location. A learned no-update option allows each cell to retain its initial features.
    \item \textbf{Observation-aware evidential occupancy completion.} We develop an occupancy-learning formulation that models observed free space and inferred occupancy through separate components. It combines spatial support with occupancy uncertainty to estimate confidence for planning.
    \item \textbf{Navigation integration and physical validation.} We integrate predicted occupancy and reliability with established local motion planners. Experiments show higher navigation success rates than the evaluated baselines in simulation and wheeled-robot trials. Additional quadruped and humanoid demonstrations illustrate deployment on other embodiments.
\end{enumerate}

\section{Related Work}
\label{sec:related_work}

\subsection{Visual Navigation and Map-Based Planning}

Learned visual navigation policies predict motion from visual observations and goal specifications. ViNT~\cite{shah2023vint} learns from diverse robot trajectories, while NoMaD~\cite{sridhar2024nomad} uses a diffusion policy for goal-directed navigation and exploration. Model-based action reannotation~\cite{hirose2025mbra} improves supervision for policy learning, and OmniVLA~\cite{hirose2026omnivla} supports goals specified through images, spatial coordinates, and language. X-NavDP~\cite{yang2026xnavdp} learns embodiment-specific behavior through robot-identity embeddings and reinforcement learning. Despite demonstrated cross-embodiment transfer, their behavior can remain sensitive to robot size, motion capabilities, and camera configurations represented in training~\cite{shah2023vint}.

Explicit traversability methods connect visual predictions to local planning. GeNIE~\cite{wang2025genie} projects single-frame traversability into a BEV cost map and identifies the absence of a maintained local map as a source of local-minimum failures.

In autonomous driving, camera-based BEV methods provide spatial representations for downstream perception and planning. Lift-Splat-Shoot~\cite{philion2020lss} and BEVFormer~\cite{li2022bevformer} combine RGB features from multiple cameras in a common BEV representation. These maps provide a spatial basis for navigation. More recent methods such as ViGT~\cite{shirokov2026vigt} predict continuous occupancy from surround-view RGB images without requiring camera calibration. The wider view helps planners identify obstacles and choose clear routes.

These methods motivate our use of a learned BEV representation. Our system supplies occupancy and reliability to motion planners. The robot footprint and traversal costs can be adjusted at planning time without changing the learned BEV predictor.

\subsection{Occupancy Prediction and Anticipation}

Occupancy anticipation addresses a further requirement for indoor navigation: inferring obstacles and free space hidden by furniture and walls. Map-based completion methods operate on partially observed top-down maps and predict occupancy in their unknown regions~\cite{sharma2023proxmap}. Geometry-guided semantic scene completion methods predict volumetric occupancy and semantic categories from RGB images using supplied camera geometry~\cite{qian2026splatssc}. Navigation-oriented methods connect hidden-space prediction with exploration and planning, including occupancy anticipation in OccAnt~\cite{ramakrishnan2020occant} and confidence-aware semantic completion in SSCNav~\cite{liang2021sscnav}.

We focus on local occupied/free completion in regions that remain unobserved across the input views. Our evidential formulation models observed free space separately from inferred occupancy and accounts for spatial support when estimating prediction confidence.

\subsection{Temporal Visual Geometry and Spatial Memory}

Spatial memory retains information from past observations to support navigation~\cite{ramakrishnan2020occant,wang2026cross}. Dense mapping has long pursued this objective through active ranging, as in LiDAR-inertial systems such as FAST-LIO2~\cite{xu2022fastlio2}. Cameras offer rich visual information without a dedicated range sensor, but recovering structure and motion from images depends on reliable correspondence and alignment. Weak texture, motion blur, and rapid viewpoint changes can challenge this estimation process~\cite{teed2021droidslam}. Learned visual SLAM improves robustness by combining learned correspondence with geometric optimization, as in DROID-SLAM~\cite{teed2021droidslam}; feed-forward models such as VGGT-Omega~\cite{wang2026vggtomega} further learn geometry across multiple images.

Using visual history for navigation requires combining observations from different viewpoints into a consistent spatial representation. Existing approaches use ego-motion or camera poses to align historical features and observations~\cite{li2022bevformer,liang2021sscnav}. Camera intrinsics and relative poses also provide geometric information for conditioning visual features~\cite{xu2026geoup}. These mechanisms depend on the supplied geometry to establish spatial relationships across views. Errors in this information can misalign historical features or maps, introducing false obstacles or distorting passage clearance and free-space connectivity.

Our work adopts the idea of learning spatial relationships across views to selectively integrate RGB history into a local BEV anchored to the latest ego frame, retaining evidence of previously observed structure. The BEV predictor decodes this history without depth or camera-pose inputs or explicit geometric warping.

\section{Methodology}
\label{sec:method}

\begin{figure*}[t]
\centering
\vspace*{5pt}
\includegraphics[width=\textwidth]{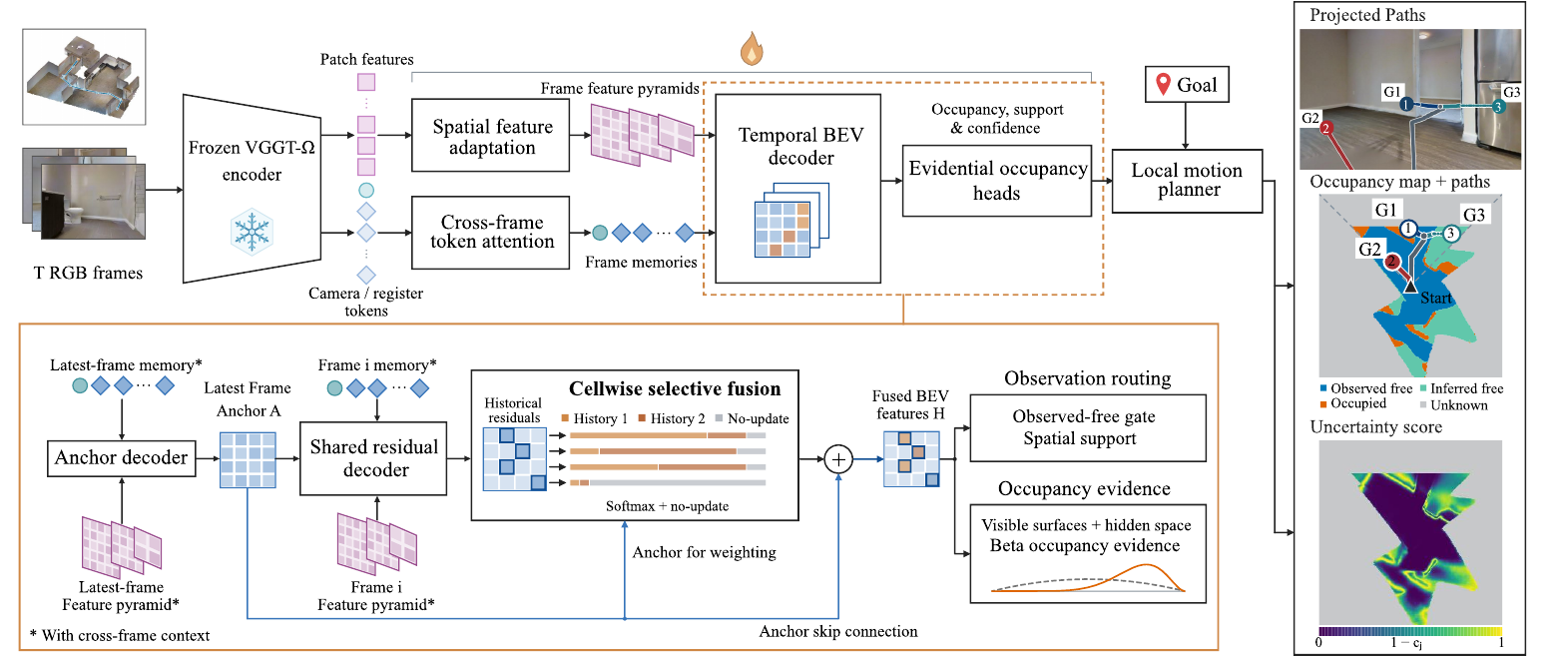}
\caption{\textbf{Framework overview.} A frozen VGGT-Omega encoder processes RGB history. Historical residuals refine a latest-frame BEV anchor through cellwise selective fusion with a learned no-update option. Observation-aware evidential heads predict occupancy, spatial support, and confidence for local planning. Right: predicted BEV with planned paths, path projections onto the latest RGB view, and uncertainty $1-c_j$.}
\label{fig:methodology}
\vspace{-1em}
\end{figure*}

\subsection{System Overview}
\label{sec:system_overview}

Given an ordered RGB window $\mathcal{I}_t=(I_{t-T+1},\ldots,I_t)$ of $T$ observations from a moving camera, we seek a local occupancy representation on a planar BEV grid $\mathcal{B}_t$ anchored to the latest ego frame. We learn the mapping
\begin{equation}
 f_{\theta}(\mathcal{I}_t)=\{(p_j,g_j,s_j,c_j)\}_{j\in\mathcal{B}_t},
 \label{eq:problem_formulation}
\end{equation}
where $\theta$ denotes the trainable parameters, and $p_j$, $g_j$, $s_j$, and $c_j$ are the occupancy probability, observed-free probability, spatial-support probability, and prediction-confidence score at cell $j$, respectively. We use a $256\times256$ grid spanning $10\times10$~m, with metric extent fixed by the training targets.

Figure~\ref{fig:methodology} summarizes the pipeline: a frozen VGGT-Omega encoder~\cite{wang2026vggtomega} extracts joint image features; a trainable decoder constructs a latest-frame BEV anchor and fuses historical residuals with cellwise weights (Sec.~\ref{sec:temporal_decoding}); observation-aware heads produce occupancy, support, and confidence (Sec.~\ref{sec:evidential_occupancy}). Complete and visibility-masked targets supervise the predictor (Sec.~\ref{sec:supervision}), whose outputs are supplied to a motion planner (Sec.~\ref{sec:planning_interface}). The BEV predictor takes RGB images without depth or camera-pose inputs.

\subsection{Selective Temporal BEV Decoding}
\label{sec:temporal_decoding}

\subsubsection{Visual features and BEV anchor}
To guide BEV queries in extracting image evidence across views, we use complementary spatial features and token context. The frozen VGGT-Omega encoder jointly processes the RGB window, with the latest image as the reference; we index this frame by $i=0$ and historical frames by $i>0$. Trainable adapters form multiscale image pyramids $F_i^\ell$ from patch features, providing spatially indexed content for sampling. Cross-frame attention over camera and register tokens produces frame-associated memories $M_i$, which condition the sampling queries with context from the window.

A learned readout of the frozen tokens produces positive frame weights $\rho_i$, normalized to mean one. These weights modulate each frame's influence when constructing the token memories: adding $\log\rho_i$ to attention logits for keys from frame $i$ favors contributions from higher-weighted frames. The same weights provide a frame-level prior for the subsequent cellwise selection of historical residuals.

For cellwise temporal selection, historical proposals should refine the same starting representation. We therefore construct a common BEV anchor in the latest ego frame. Let $Q$ denote the learned BEV queries with positional encodings, and let $\mathcal{R}_i(X)=\operatorname{Attn}(X,M_i)$ denote query attention to frame $i$'s token memory. The anchor is
\begin{equation}
 A=\mathcal{D}_0\!\left(Q+\mathcal{R}_0(Q),\{F_0^{\ell}\}_{\ell}\right).
 \label{eq:anchor}
\end{equation}
The anchor decoder samples the latest image pyramid using deformable cross-attention and learns image-to-BEV correspondence through BEV supervision without explicit camera projection. Subsequent historical branches propose residual refinements from this same anchor, allowing their contributions to be selected at each BEV location.

\subsubsection{Historical residual proposals}
A shared decoder generates historical residual proposals from the anchor $A$. Each proposal is conditioned on the corresponding frame's image pyramid and token context. The resulting residual is given by
\begin{equation}
 C_i=\mathcal{R}_i(A),\qquad
 \Delta_i=\mathcal{D}_{h}\!\left(A,\{F_i^{\ell}\}_{\ell},C_i\right)-A.
 \label{eq:proposal}
\end{equation}
The token context $C_i$ conditions the queries used to sample the historical image pyramid, producing a feature residual $\Delta_i$ at each anchor cell. Residual-output projections are zero-initialized to preserve the anchor at initialization.

\subsubsection{Spatially selective fusion}
Historical views provide different evidence at different map locations, motivating cellwise control over both their relative contributions and the strength of their updates. We include a zero-residual candidate so that fusion can also control the total strength of the historical update.

For $T>1$, a learned scorer combines the anchor $A_j$, residual $\Delta_{ij}$, and token context $C_{ij}$ at each cell. Adding $\log\rho_i$ and a learned frame-age bias gives the historical score $e_{ij}$. A no-update candidate has score $e_{\varnothing j}=f_{\varnothing}(A_j)+\log(T-1)+\log[\pi_0/(1-\pi_0)]$, where $f_{\varnothing}$ is learned. The candidate weights are
\begin{equation}
 w_{ij}=\frac{\exp(e_{ij})}{\sum_{k\in\mathcal{J}}\exp(e_{kj})},
 \qquad \mathcal{J}=\{\varnothing,1,\ldots,T-1\}.
 \label{eq:temporal_weights}
\end{equation}
The learned frame weight $\rho_i$ provides a spatially shared prior, while $w_{ij}$ determines that frame's contribution separately at each BEV cell.

With uniform initial frame weights and zero learned scoring terms, the bias gives an initial no-update probability of $\pi_0=0.9$. The fused representation is
\begin{equation}
 H_j=A_j+\sum_{i=1}^{T-1}w_{ij}\Delta_{ij}.
 \label{eq:fusion}
\end{equation}
A higher no-update weight leaves less weight for historical residuals, reducing their combined correction to the anchor. The decoder maintains no recurrent BEV state and returns $H=A$ when $T=1$. This design allows historical evidence to refine individual BEV cells while retaining the anchor as the default representation.

\subsection{Observation-Aware Evidential Occupancy}
\label{sec:evidential_occupancy}

From the fused representation $H$, the routing branch predicts observed-free probability $g_j$ and spatial-support probability $s_j$. The former estimates whether a cell is free and visible in any input frame. Spatial support covers the union of the input camera footprints wherever complete labels are valid, including occluded cells. The evidence branch models visible occupied surfaces and hidden free or occupied cells. This separation distinguishes observed free space from inferred free space, even when both yield a low occupancy probability.

Two logits parameterize a Beta distribution~\cite{sensoy2018evidential} over occupancy probability, with $\alpha_j=1+\operatorname{softplus}(z_j^o)$, $\beta_j=1+\operatorname{softplus}(z_j^f)$, and mean $\mu_j=\alpha_j/(\alpha_j+\beta_j)$. Conditional on the observed-free component, occupancy is modeled as zero. Mixing this component with the Beta distribution using weights $g_j$ and $1-g_j$ gives
\begin{equation}
 p_j=(1-g_j)\mu_j.
 \label{eq:occupancy}
\end{equation}
With Beta variance $v_j=\mu_j(1-\mu_j)/(\alpha_j+\beta_j+1)$, the mixture variance and composite confidence are
\begin{align}
 V_j&=(1-g_j)v_j+g_j(1-g_j)\mu_j^2,\label{eq:mixture_variance}\\
 c_j&=s_j\,|2p_j-1|\,\operatorname{clip}(1-4V_j,0,1).
 \label{eq:confidence}
\end{align}
The mixture variance captures spread within the evidence distribution and disagreement between components. High confidence requires strong spatial support, a clear occupied/free preference, and low mixture variance. These factors reduce confidence where occupancy predictions are ambiguous or variable, while $s_j$ suppresses confidence in cells with low predicted support even when one class is strongly favored. The resulting confidence is a constructed score rather than a calibrated collision probability.

\subsection{Supervision and Optimization}
\label{sec:supervision}

Complete and visibility-masked targets are expressed in the latest ego frame. We denote spatial support by $U$, observed free space by $O\subseteq U$, and the remaining region by $G=U\setminus O$. These regions supervise spatial support $s_j$, observed-free routing $g_j$, and occupancy evidence, respectively.

For each cell $j\in G$, we minimize the Bernoulli negative log-likelihood averaged under the predicted Beta distribution, with $y_j=1$ denoting an occupied cell:
\begin{equation}
 \ell_j^E=\psi(\alpha_j+\beta_j)
 -y_j\psi(\alpha_j)-(1-y_j)\psi(\beta_j),
 \label{eq:edl_loss}
\end{equation}
where $\psi$ is the digamma function. The occupancy objective is $\mathcal{L}_{\mathrm{occ}}=\langle\ell^E+\lambda_K\ell^K\rangle_w$. The KL regularizer suppresses evidence for the incorrect class. It compares the predicted Beta distribution to the uniform $\operatorname{Beta}(1,1)$ prior after resetting the correct-class parameter to one. To balance regions of unequal area, we average the occupancy loss separately over hidden free cells, visible occupied surfaces, and hidden occupied cells in each sample. Their means are then combined using relative weights normalized over the regions present.

Class-balanced BCE defines $\mathcal{L}_{\mathrm{gate}}$ using positives in $O$ and negatives in $G$, and $\mathcal{L}_{\mathrm{support}}$ using positives in $U$ and negatives outside $U$. The resulting map objective jointly supervises observed-free routing, occupancy evidence, and spatial coverage:
\begin{equation}
 \mathcal{L}_{\mathrm{BEV}}=
 \mathcal{L}_{\mathrm{gate}}+\mathcal{L}_{\mathrm{occ}}
 +\lambda_B\mathcal{L}_{\mathrm{support}}.
 \label{eq:bev_loss}
\end{equation}
We use $\lambda_B=0.5$ and a maximum regularization weight $\lambda_K=0.005$. Hidden-occupied supervision and wrong-class evidence regularization are introduced gradually during training.

\subsection{Inference and Planning}
\label{sec:planning_interface}

The planner uses the predicted BEV and a metric point goal, both expressed in the latest ego frame. Supported occupied cells are inflated according to the robot footprint and safety margin to form hard collision constraints. Traversal costs combine occupancy $p_j$, obstacle clearance, and distribution reliability $\kappa_j=s_j\operatorname{clip}(1-4V_j,0,1)$. Within supported traversable regions, lower reliability incurs a larger penalty. Unsupported cells retain their unknown labels and incur finite traversal penalties, subject to the collision constraints.

When the goal lies outside the map or the predicted reachable region, the planner selects a local subgoal within that region. Only a short path segment is executed, with an additional limit on travel through unknown space. The next RGB observation updates the BEV before replanning.

\section{Experiments}
\label{sec:experiments}

\begin{figure*}[t]
\centering
\includegraphics[width=\textwidth]{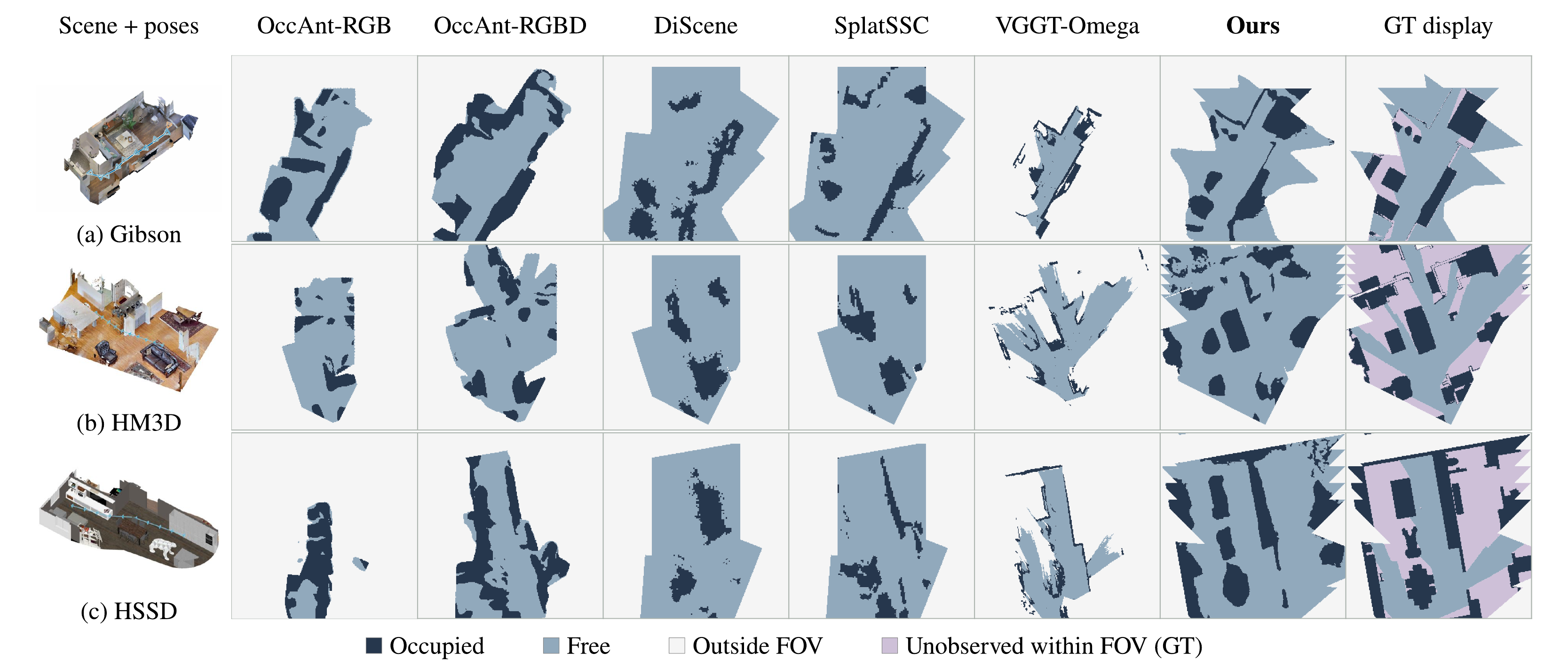}
\caption{\textbf{Representative BEV completion results.} Maps from image history are shown on a common grid for Gibson, HM3D, and HSSD scenes. Purple in GT highlights regions within the input camera-footprint union that remain unobserved across all views and require completion. Off-white denotes regions without an occupancy label.}
\label{fig:unseen_qualitative}
\vspace{-1em}
\end{figure*}

In this section, we evaluate the proposed representation from map prediction to navigation. We present our model training procedure in Sec.~\ref{sec:experimental_setup} and the comparison with baselines in Sec.~\ref{sec:occupancy_experiments}. Section~\ref{sec:inference_ablation} presents the temporal fusion ablation. Sections~\ref{sec:navigation_experiments} and~\ref{sec:robot_experiments} report simulation navigation and physical-robot evaluation, respectively.

\subsection{Data Collection and Training}
\label{sec:experimental_setup}

\subsubsection{Data collection}
We collect RGB sequences along sampled navigable paths in Habitat environments from HM3D~\cite{ramakrishnan2021hm3d}, HSSD~\cite{khanna2024hssd}, and Matterport3D~\cite{chang2017mp3d}, varying camera height, field of view, and motion speed. The indoor corpus contains approximately 535k training sessions from 1k scenes and 26k validation sessions from 54 disjoint scenes. For each session, we generate complete and visibility-masked BEV targets in the latest ego frame, following the target definitions in Sec.~\ref{sec:supervision}.

\subsubsection{Training configuration}
We freeze the pretrained VGGT-Omega encoder and optimize the BEV adapters, temporal decoder, and prediction heads. The decoder accepts variable-length RGB sequences. During training, BoQ-based visual similarity guides frame selection to reduce visual redundancy, with sequence lengths centered on ten frames. Predictions use a $256\times256$ BEV grid spanning $10\times10$~m. Hidden-occupied supervision and wrong-class evidence regularization are introduced gradually during training.

Gibson~\cite{xia2018gibson} and ProcTHOR~\cite{deitke2022procthor} are unseen datasets used only for evaluation and are entirely excluded from our BEV training data.

\subsection{Occupancy Prediction under Partial Observability}
\label{sec:occupancy_experiments}

\subsubsection{Evaluation settings}
We evaluate 1,200 matched routes from 101 scenes in HM3D, HSSD, and Gibson, with 400 routes per dataset. Scenes used for our BEV training are excluded. Single-image prediction uses the final image of each ten-frame route on a forward-facing $3.25\times3.25$~m grid with 5~cm cells. History mapping uses all ten images on a $256\times256$ grid spanning $10\times10$~m. Both settings share the same final ego frame, and evaluation covers valid labels within the corresponding camera-footprint union.

We compare against DiScene~\cite{discene2026}, SplatSSC~\cite{qian2026splatssc}, VGGT-Omega~\cite{wang2026vggtomega}, OccAnt-RGB, and OccAnt-RGBD~\cite{ramakrishnan2020occant}. DiScene and SplatSSC receive camera calibration; their volumetric predictions are projected into BEV and accumulated across views using ground-truth poses. For OccAnt, we bypass learned pose estimation and use ground-truth poses for inter-frame accumulation, removing odometry errors from the comparison. OccAnt-RGBD additionally receives sensor depth. VGGT-Omega maps are projected from predicted geometry and cameras using camera-height calibration. Our predictor receives only RGB images, without camera intrinsics or inter-frame extrinsics.

\subsubsection{Metrics and filtering}
Following Sec.~\ref{sec:supervision}, $O$ denotes observed free space and $G=U\setminus O$ the remaining supported region. We report free-space IoU in $O$ and occupied/free mIoU in $G$ and $U$. Unknown predictions count as errors in $O$ and are excluded from $G$. Thus, mIoU$_G$ measures accuracy conditional on a known prediction, while mIoU$_U$ combines the regional counts under the same policy. We also report $\mathrm{Cov}_G$, the fraction of $G$ assigned an occupied or free label.

For our model, dense prediction labels every evaluated cell as occupied when $p_j\geq0.5$ and free otherwise. Filtered prediction applies the same labeling rule only to cells satisfying $g_j\geq0.5$ or $c_j\geq0.4$; the remaining cells are marked unknown. OccAnt assigns occupied/free labels only to cells with explored probability $>0.6$; the remaining cells are marked unknown.

\begin{table}[t]
\vspace*{5pt}
\centering
\caption{Occupancy Prediction Results}
\label{tab:occupancy_single}
\label{tab:occupancy_history}
\setlength{\tabcolsep}{3pt}
\begin{tabular}{@{}lrrrr@{}}
\toprule
Method & mIoU$_U$ & IoU$_O$ & mIoU$_G$ & $\mathrm{Cov}_G$ \\
\midrule
\multicolumn{5}{l}{Single image: $3.25\times3.25$~m}\\
\midrule
DiScene & 43.21 & 97.28 & 32.02 & 100.00 \\
SplatSSC & 40.58 & 94.97 & 31.11 & 100.00 \\
VGGT-Omega & 42.62 & 53.12 & 44.26 & 13.06 \\
OccAnt-RGB & 32.40 & 44.88 & 33.88 & 39.14 \\
OccAnt-RGBD & 63.96 & 81.57 & \textbf{55.16} & 56.32 \\
\textbf{Ours (dense)} & 57.32 & \textbf{97.54} & 48.29 & 100.00 \\
\textbf{Ours (filtered)} & \textbf{64.72} & 96.98 & 52.79 & 50.84 \\
\midrule
\multicolumn{5}{l}{Ten-image history: $10\times10$~m}\\
\midrule
DiScene & 39.74 & 80.36 & 32.68 & 64.76 \\
SplatSSC & 37.71 & 75.97 & 32.12 & 64.76 \\
VGGT-Omega & 40.61 & 55.25 & 38.22 & 11.86 \\
OccAnt-RGB & 31.10 & 47.15 & 31.27 & 22.26 \\
OccAnt-RGBD & 55.82 & 80.14 & 46.72 & 42.51 \\
\textbf{Ours (dense)} & 55.27 & \textbf{95.67} & 47.91 & 100.00 \\
\textbf{Ours (filtered)} & \textbf{68.32} & 89.58 & \textbf{61.17} & 35.03 \\
\bottomrule
\end{tabular}
\vspace{\dimexpr-1.5em-5pt\relax}
\end{table}

\subsubsection{Occupancy completion}
Region $G$ includes visible obstacle surfaces and regions that remain occluded across the input views, where free space and obstacles must be inferred. Figure~\ref{fig:unseen_qualitative} shows that VGGT-Omega leaves parts of the surrounding layout unknown, while our model predicts both obstacles and free space in these regions. For single-image prediction, our model achieves the highest reported mIoU$_G$ among RGB-based methods, exceeding the strongest RGB baseline, VGGT-Omega, by 4.03 percentage points with dense prediction and 8.53 points after filtering. Dense prediction covers all of $G$, compared with 13.06\% coverage for VGGT-Omega. Our filtered predictions also achieve mIoU$_G$ within 2.37 percentage points of depth-assisted OccAnt-RGBD, with coverage 5.48 percentage points lower. In history mapping, our dense predictions outperform all baselines, exceeding OccAnt-RGBD by 1.19 points while providing full coverage. Filtering increases this margin over the strongest baseline, OccAnt-RGBD, to 14.45 percentage points in mIoU$_G$ at 35.03\% coverage of $G$. These results support dense occupancy prediction beyond observed free space and a selective alternative with higher conditional accuracy at reduced coverage.

\subsubsection{Temporal BEV mapping}
Temporal mapping errors can produce fragmented occupied regions and displaced or duplicated boundaries, such as those seen in several baseline maps in Fig.~\ref{fig:unseen_qualitative}. Such errors can falsely block passages or omit obstacles, leading to detours or increased collision risk. Our model predicts a unified BEV from RGB history in one forward pass, while the baselines accumulate predictions using predicted or ground-truth camera extrinsics. Our filtered predictions achieve the highest reported mIoU$_U$ in both settings. Their advantage over the strongest baseline increases from 0.76 percentage points for single-image prediction to 12.50 points for history mapping. Dense prediction also shows a larger advantage over the strongest RGB baseline in the history setting. Its mIoU$_U$ decreases by only 2.05 percentage points, compared with 8.14 points for OccAnt-RGBD. These results indicate that our model effectively retains and integrates information from past RGB observations to construct a more complete map.

\subsection{Temporal Fusion Ablation}
\label{sec:inference_ablation}

We examine temporal fusion on 190 examples, comprising 11 HM3D and 179 ProcTHOR scenes excluded from training. All variants use the same checkpoint. Anchor-only prediction omits historical residual updates while retaining historical context from joint encoding. Equal history weighting preserves the learned no-update weight at each cell and distributes the remaining weight equally among historical residuals.

All variants produce dense occupancy predictions (Table~\ref{tab:temporal_fusion_ablation}). Recovery denotes cells visible in earlier observations but not in the latest view; Hidden denotes cells unobserved across all input views.

Full selective fusion raises overall mIoU from 47.59\% for anchor-only prediction to 58.43\%. Compared with equal history weighting, it improves mIoU by 1.90 percentage points overall, 2.35 in Recovery, and 2.25 in Hidden. These results demonstrate the contribution of learned historical weighting to temporal recovery and occupancy completion within the trained model.

\begin{table}[!ht]
\centering
\caption{Temporal fusion ablation.}
\label{tab:temporal_fusion_ablation}
\setlength{\tabcolsep}{3pt}
\begin{tabular}{@{}lrrr@{}}
\toprule
Fusion variant (mIoU, \%) & Overall & Recovery & Hidden \\
\midrule
No history fusion & 47.59 & 45.79 & 37.78 \\
Equal-weight fusion & 56.53 & 46.60 & 48.01 \\
Full selective fusion & \textbf{58.43} & \textbf{48.95} & \textbf{50.26} \\
\bottomrule
\end{tabular}
\vspace{-1em}
\end{table}

\subsection{Simulation Navigation}
\label{sec:navigation_experiments}

\subsubsection{Comparisons and metrics}
We evaluate simulated wheeled-robot navigation against LoGoNav~\cite{hirose2025mbra}, OmniVLA~\cite{hirose2026omnivla}, X-NavDP~\cite{yang2026xnavdp}, and OccAnt-RGBD~\cite{ramakrishnan2020occant} in both the sequential and hidden-goal benchmarks. Ours and OccAnt use Dijkstra with a shared path executor. X-NavDP and OccAnt-RGBD receive sensor depth; the remaining methods use RGB only. OccAnt accumulates map predictions using ground-truth odometry. Goal acceptance requires a collision-free connection from the robot to the goal.

For $N$ episodes, let $\zeta_n\in\{0,1\}$, $P_n$, and $L_n$ denote success, executed distance, and precomputed ground-truth reference path length, respectively. We compute $L_n$ by summing edge lengths along the precomputed reference path in the ground-truth navigability graph. Goal tolerance and continuous motion allow successful trajectories to be shorter than this discrete reference, yielding SCPR below one. We report success rate (SR), SPL~\cite{anderson2018evaluation}, and success-conditioned path ratio (SCPR):
\begin{gather}
\mathrm{SR}=\frac{1}{N}\sum_n\zeta_n,
\qquad
\mathrm{SCPR}=
\frac{\sum_n\zeta_n P_n/L_n}{\sum_n\zeta_n},\nonumber\displaybreak[1]\\
\mathrm{SPL}=\frac{1}{N}\sum_n\zeta_n
\frac{L_n}{\max(L_n,P_n)}.
\label{eq:navigation_metrics}
\end{gather}
Failures contribute zero to SPL. SCPR is averaged over successful episodes; methods with no successful episodes are excluded from the SCPR comparison.

\begin{figure}[!tb]
\centering
\includegraphics[width=\columnwidth]{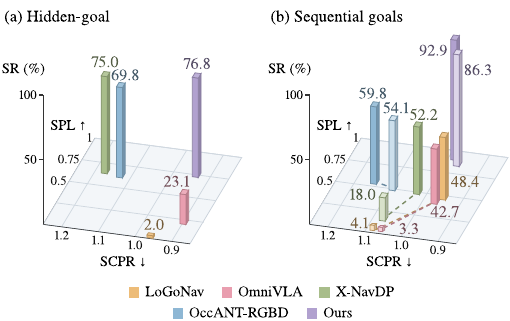}
\caption{Simulation navigation results for (a) hidden-goal and (b) sequential two-goal tasks. Bar height indicates SR; ground-plane coordinates indicate SPL and SCPR. In (b), darker and lighter bars represent reaching the first goal and both goals, respectively.}
\label{fig:navigation}
\vspace{-1em}
\end{figure}

\subsubsection{Sequential two-goal navigation}
The sequential two-goal navigation benchmark comprises 1,000 shared routes across 308 scenes excluded from training, with 250 routes each from Gibson, HSSD, HM3D, and ProcTHOR. The second destination is revealed only after the first is reached (Fig.~\ref{fig:navigation_comparison}). The current waypoint comes from a precomputed ground-truth route and is expressed relative to the robot at planning time. Ground-truth occupancy maps and future waypoints are not provided. Success requires reaching both destinations in order without collision.

For the first goal, Ours achieves an SR of 92.9\% and an SPL of 0.922, exceeding the strongest remaining baseline, OccAnt-RGBD, by 33.1 percentage points in SR and 0.370 in SPL. Ours and OccAnt account for the robot's footprint and clearance. Both show stronger performance in guiding the robot along efficient, collision-free paths.

After the goal changes, Ours maintains 86.3\% SR and 0.860 SPL. Its SR decreases by only 6.6 percentage points, compared with decreases of 5.7 to 44.3 points for the baselines. Figure~\ref{fig:navigation_comparison} shows two challenging navigation cases where only our method reaches both goals. OccAnt fails due to a missed obstacle boundary in Case 1 and overestimated obstacle extent in Case 2. Our maps more accurately represent the obstacles and surrounding free space, allowing the planner to find routes to both goals.

Our method achieves a first-goal SCPR of 0.983, comparable to LoGoNav's 0.951. It completes nearly twice as many episodes as LoGoNav. It also succeeds in challenging scenarios with similar path efficiency. For the complete trajectory, our SCPR of 0.967 is the lowest among all evaluated methods. Only our method and OccAnt-RGBD show lower SCPR over the complete trajectory, while the other baselines show increases. This pattern provides the evidence that retained historical observations can help guide more efficient path planning as navigation continues.

\subsubsection{Navigation toward initially hidden destinations}
To assess navigation beyond initially observed free space, we construct 1,000 tasks in 503 existing ProcTHOR houses. Each task specifies a reachable goal 1--3~m from the start, occluded by an object visible in the initial image. The relative goal is supplied once, and navigation proceeds with subsequent observations without intermediate waypoints.

Our method achieves the highest SR and SPL among all evaluated methods, with a substantial advantage over the other RGB-based approaches (Fig.~\ref{fig:navigation}). It also exceeds the depth-assisted baselines on both metrics, demonstrating effective navigation toward initially hidden destinations using RGB observations alone.

Our method achieves substantially lower SCPR than X-NavDP and OccAnt-RGBD, the two strongest baselines in SR and SPL. OmniVLA achieves a lower SCPR than ours, but succeeds in far fewer episodes. Our method combines high navigation success with efficient paths toward initially hidden goals.

\begingroup
\setlength{\intextsep}{16pt}
\begin{figure}[!ht]
\centering
\includegraphics[width=\columnwidth]{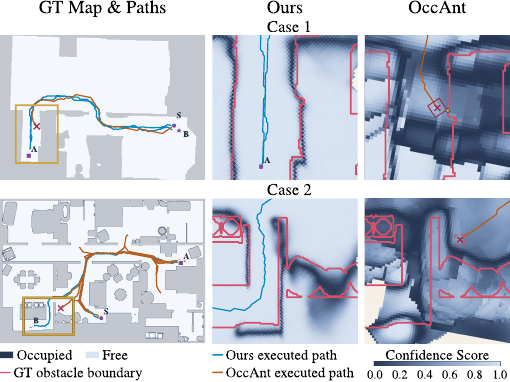}
\caption{Sequential two-goal navigation comparing Ours and OccAnt. Yellow boxes mark the enlarged regions; S denotes the start and A/B the goals.}
\label{fig:navigation_comparison}
\vspace{-1em}
\end{figure}
\endgroup

\makeatletter
\@startsection{subsection}{2}{\z@}{1.5ex}
  {0.7ex}{\normalfont\normalsize\itshape}
  {Physical-Robot Evaluation}
\makeatother
\label{sec:robot_experiments}

\begin{figure}[t]
\vspace*{3pt}
\centering
\includegraphics[width=\columnwidth]{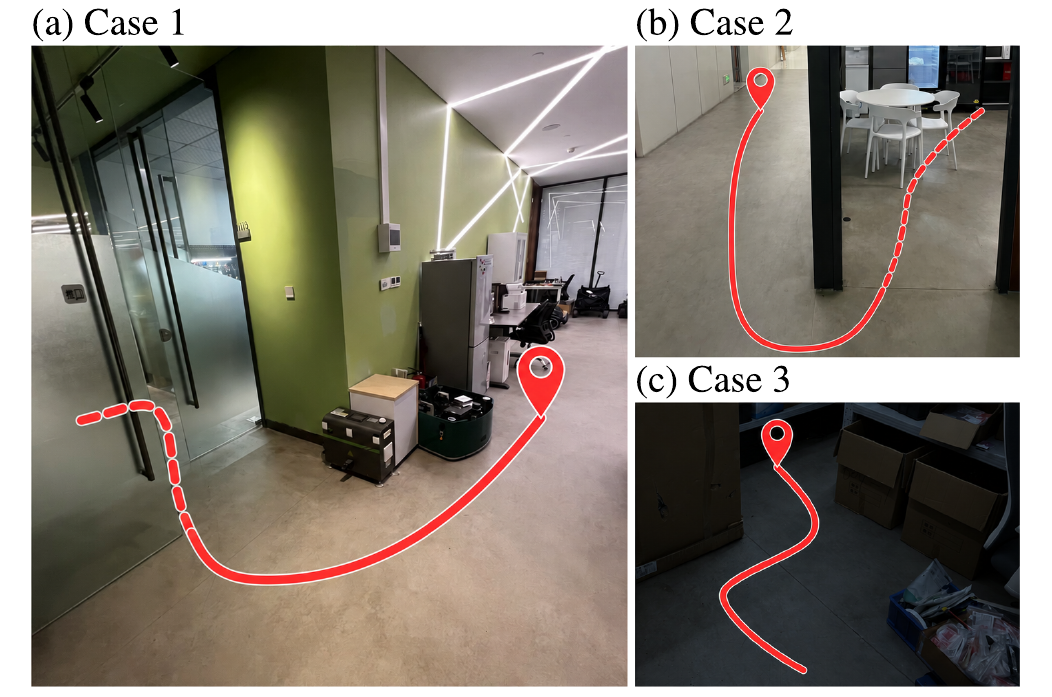}
\caption{Physical-robot navigation examples in three indoor scenes, with routes overlaid in red.}
\label{fig:real_robot_examples}
\vspace{-1em}
\end{figure}

We deploy our framework on a Mecanum-wheeled robot with an Intel RealSense D435 camera. Our BEV predictor processes temporal RGB observations, and Dijkstra search plans paths on the predicted occupancy map. We evaluate Ours, X-NavDP, and OmniVLA in 20 scenes, with three matched trials per scene. Success requires reaching within 30~cm of the goal. A trial fails if the robot collides with an obstacle or the planner cannot find a feasible path. Figure~\ref{fig:real_robot_examples} shows representative scenes from our physical-robot experiments. The frosted-glass doorway in (a) restricts visibility into the adjacent space, while (b) requires planning through a doorway toward a goal outside the room. Very low illumination in (c) reduces the visual detail available for perception.

Our method achieves a success rate of 73.3\%, compared with 33.3\% for X-NavDP and 26.7\% for OmniVLA. It completes more than twice as many trials as the strongest evaluated baseline, X-NavDP. These results demonstrate the practical value of integrating temporal BEV mapping and occupancy completion with conventional path planning, enabling successful physical navigation using a spatial representation predicted from RGB observations. We further demonstrate navigation on humanoid and quadruped robots. These qualitative demonstrations extend the framework's deployment to platforms with different locomotion capabilities.


\section{Conclusion}
\label{sec:conclusion}

We presented an RGB-based local navigation framework combining selective temporal BEV fusion with observation-aware evidential occupancy completion. The predictor integrates historical observations and infers occluded obstacles and free space without depth or camera-pose inputs. It distinguishes observed free space from inferred occupancy and provides confidence for local planning. The integrated system achieves the highest success rate and SPL in both simulated navigation benchmarks and the highest success rate in matched wheeled-robot trials. Additional qualitative demonstrations extend deployment to humanoid and quadruped robots. These results establish BEV maps that integrate RGB history and complete occluded regions as an effective interface between perception and local planning under partial observability.

\par
\setlength{\textfloatsep}{18.71pt}
\setlength{\parskip}{0pt}
\makeatletter
\def\section{\@startsection{section}{1}{\z@}{5.79pt}
  {2.83pt}{\normalfont\normalsize\centering\scshape}}
\patchcmd{\thebibliography}
  {\vskip 0.3\baselineskip plus 0.1\baselineskip minus 0.1\baselineskip}
  {\vskip 0.3\baselineskip}
  {}{\PackageError{EchoNav}{Could not fix References spacing}{Check ieeeconf.cls.}}
\makeatother
\microtypesetup{protrusion=false}
\setlength{\emergencystretch}{1em}
\bibliographystyle{IEEEtran}
\bibliography{references}
\end{document}